\documentclass[lettersize,journal]{IEEEtran}
\usepackage{graphicx}
\usepackage{amssymb}
\usepackage{amsthm}

\usepackage{amsmath}

\usepackage{setspace}
\usepackage[utf8]{inputenc}
\usepackage[noend]{algpseudocode}

\usepackage{comment}
\usepackage{enumitem} 
\usepackage{tablefootnote}
\usepackage{caption}
\usepackage{graphicx}
\usepackage{booktabs}
\usepackage{amsfonts}
\usepackage{amsthm}
\usepackage{amsmath,bm}
\usepackage{amssymb}
\usepackage{mathtools}
\usepackage{subfig}
\usepackage{multirow}
\usepackage{bbm}
\usepackage{hyperref}

\newtheorem{definition}{Definition}
\usepackage{array}
\usepackage{tikz}
\usetikzlibrary{shapes, matrix, positioning}
\usepackage[colorinlistoftodos]{todonotes}
\usepackage{comment}

\usepackage{mathrsfs}

\usepackage{cuted}

\usepackage{dsfont}

\usepackage{epstopdf}    
\usepackage{enumitem}  

\usepackage{diagbox}

\usepackage[linesnumbered,ruled]{algorithm2e}

\DeclareMathOperator{\EX}{\mathbb{E}}
\begin{document}

\title{XRMDN: An Extended Recurrent Mixture Density Network for Short-Term Probabilistic Rider Demand Forecasting with High Volatility}

\author{Xiaoming~Li $^{1}$, 
        Hubert Normandin-Taillon $^{2}$,
        Chun~Wang $^{3}$,
        Xiao~Huang $^{4}$
\thanks{$^{1}$Xiaoming Li is with Concordia Institute for Information Systems Engineering (CIISE), Concordia University, Montréal, QC H3G 1M8, Canada {\tt\small xiaoming.li@concordia.ca}}%
\thanks{$^{2}$Hubert Normandin-Taillon is with the Department of Computer Science and Software Engineering (CSSE), Concordia University, Montréal, QC H3G 1M8, Canada
        {\tt\small hubert.normandin-taillon@mail.concordia.ca}}%
\thanks{$^{3}$Chun Wang is with Concordia Institute for Information Systems Engineering (CIISE), Concordia University, Montréal, QC H3G 1M8, Canada
        {\tt\small chun.wang@concordia.ca}}%
\thanks{$^{4}$Xiao Huang is with Concordia John Molson School of Business (JMSB), Concordia University, Montréal, QC H3G 1M8, Canada
        {\tt\small xiao.huang@concordia.ca}}%
}



\maketitle

\begin{abstract}

In the realm of Mobility-on-Demand (MoD) systems, the forecasting of rider demand is a cornerstone for operational decision-making and system optimization. Traditional forecasting methodologies primarily yield point estimates, thereby neglecting the inherent uncertainty within demand projections. Moreover, MoD demand levels are profoundly influenced by both endogenous and exogenous factors, leading to high and dynamic volatility. This volatility significantly undermines the efficacy of conventional time series forecasting methods. In response, we propose an Extended Recurrent Mixture Density Network (XRMDN), a novel deep learning framework engineered to address these challenges. XRMDN leverages a sophisticated architecture to process demand residuals and variance through correlated modules, allowing for the flexible incorporation of endogenous and exogenous data. This architecture, featuring recurrent connections within the weight, mean, and variance neural networks, adeptly captures demand trends, thus significantly enhancing forecasting precision, particularly in high-volatility scenarios. Our comprehensive experimental analysis, utilizing real-world MoD datasets, demonstrates that XRMDN surpasses the existing benchmark models across various metrics, notably excelling in high-demand volatility contexts. This advancement in probabilistic demand forecasting marks a significant contribution to the field, offering a robust tool for enhancing operational efficiency and customer satisfaction in MoD systems.

\end{abstract}


\section{Introduction}

\IEEEPARstart{I}{n} the transformative landscape of urban mobility, Mobility-on-Demand (MoD) systems have emerged as pivotal to reshaping how city dwellers navigate their environments, placing demand forecasting at the heart of operational efficiency and customer satisfaction. Accurate predictions are essential for optimizing fleet operations and enhancing user experiences. Particularly in MoD systems such as car-sharing and ride-hailing services, the ability to accurately forecast short-term demand is critical due to the unpredictable and highly volatile nature of user requests. This \textit{volatility} can lead to significant operational challenges, as abrupt increases in demand can overwhelm unprepared systems, disrupting the balance between supply and demand~\cite{yan2020dynamic}.

The demand for MoD systems in the short term exhibits distinct characteristics that are crucial for time series models to accurately forecast. First, exogenous factors, such as spatial and meteorological features, exert a substantial influence on demand levels. Secondly, this demand is prone to significant and unpredictable fluctuations~\cite{chandakas2020demand}, largely due to its reliance on temporal variables such as the distinction between weekdays and weekends, as well as peak versus off-peak periods~\cite{ding2017using}. The prevailing methodologies for forecasting MoD demand bifurcate into three principal frameworks: statistical, machine learning (ML), and deep learning (DL) models. The majority of statistical approaches hinge on the presumption that current demand is linearly correlated with historical sequences of demand, applying linear regression techniques for predictive analyses. This presumption, however, fails to encapsulate the intrinsic non-linearity of short-term demand, particularly in instances of high volatility. In contrast, ML and DL models are proficient in assimilating non-linear patterns from historical data by integrating both internal and external informational cues. Despite of this, such models predominantly yield point forecasts or expected demand results, which lack consideration for the \textit{uncertainty} inherent in future demand levels. This omission may hinder the capacity for effective strategic planning and decision-making in the face of dynamic demand scenarios~\cite{chen2023probabilistic,fatema2023probabilistic}.

The Recurrent Mixture Density Networks (RMDN) model~\cite{schittenkopf_forecasting_2000} is a probabilistic forecasting model that approximates the distribution of time series forecasts. However, RMDN has limitations when being applied to short-term demand forecasting in MoD systems. first, it does not encode exogenous information, such as spatio-temporal and meteorological features. second, in RMDN, the recurrent connections are only applied in the variance neural network while neglecting the demand auto-correlations in the weight and mean neural networks. This may lead RMDN to fail to utilize the latent information in the MoD demand sequence. 

To address the challenges, we propose a novel deep learning structure by extending the classic RMDN to forecast the short-term probabilistic demand with high volatility in MoD systems, in which three correlated recurrent neural networks are combined via the demand residual to capture the trends and dependencies of rider demand. The proposed end-to-end solution redefines the demand probabilistic forecasting paradigm by approximating a Gaussian mixture model whose parameters are obtained from three correlated recurrent neural networks. Furthermore, we conduct comprehensive experiments comparing the performance of the proposed architecture to the benchmark time series forecasting approaches, including statistical, machine learning, and deep learning models. To the best of our knowledge, this is the first deep learning-based work to study short-term probabilistic demand forecasting while considering the rider demand with high volatility in MoD systems.

The rest of the paper is organized as follows. In Section\ref{section:literature}, we conduct a comprehensive literature review on the time series forecasting approaches in MoD systems. In Section~\ref{section:approach}, we first briefly describe the problem setting of probabilistic demand forecasting in MoD systems, then present the proposed \textit{Extended Recurrent Mixture Density Network} (XRMDN). We conduct a numeric study using two real-world datasets to compare the XRMDN model to the three types of benchmark time series forecasting models in Section~\ref{section:experiment}. Finally, we summarize the work and discuss the future research directions in Section~\ref{section:conclusion}.

\section{Literature Review}\label{section:literature}
There are three major streams of short-term demand forecasting approaches in MoD systems. Namely, statistical models, machine learning models, and the dominant wave of deep learning models. 

Most of the statistical models are based on the Box-Jenkins method~\cite{box2015time}. In particular, the linear model AutoRegession Integrated Moving Average (ARIMA)~\cite{rodrigues2020spatiotemporal}, and its variants~\cite{chandakas2020demand} are widely applied in short-term demand forecasting, which leverages previous time-dependent values and residual parts. However, these models are limited to be applied in demand forecasting when the demands are highly volatile due to the models' linearity property. In addition, considering the time-dependent variation (e.g., demand volatility), statistical volatility models have been applied for short-term demand forecasting. In particular, AutoRegssive Conditional Heteroskedasticity (ARCH)~\cite{engle1982} and Generalized AutoRegressive Conditional Heteroskedasticity (GARCH)~\cite{bollerslev1986} are widely applied. ARCH was coined by Engle for financial data, which assumes that the time series volatility is predictable. More recently, the mean of the GARCH model has been modeled as an ARIMA process. The resulting model, known as ARIMA-GARCH, has been applied to short-term demand forecasting with heteroskedastic demand in MoD applications. For example, short-term ridership forecasting in the subway~\cite{ding2017using}, short-term traffic flow prediction~\cite{chen2011short, zhang2014hybrid}. Benefiting from the combination of the ARIMA and GARCH model, the ARIMA-GARCH can capture the conditional mean and conditional variance of MoD demand. However, the residual in GARCH is assumed to follow a Gaussian distribution that may not accurately reflect the correct volatility in real MoD demand. In addition, most of the statistical models do not encode the exogenous information that impacts the MoD demand forecasting results.

Machine learning models, such as random forest~\cite{breiman2001random}, gradient boosting, k-nearest neighbor (KNN), and support vector regression (SVR), are also applied in short-term demand forecasting issues. For example, Li et al.~\cite{li2011hunting} design an L-1 SVM approach to forecast the taxi demand where some interesting hidden facts can be revealed. Toque et al.~\cite{toque2017short} propose two multi-modal transport passenger flows forecasting models based on random forest. Guo et al.~\cite{guo2019short} combine SVR and LSTM as the two-stage time series forecasting method to predict abnormal passenger flow. The hybrid model leverages historical data and real-time data to forecast the short-term abnormal passenger flow. Considering the time series data abruptly varied or exhibited wide fluctuations, Chang et al.~\cite{chang2012dynamic} address the short-term prediction by proposing a dynamic multi-interval traffic volume prediction model, which is based on KNN regression. As the ensemble learning approach, gradient boosting models are also applied in time series demand prediction in MoD applications, such as taxi demand prediction~\cite{vanichrujee2018taxi} and bike-sharing demand prediction~\cite{ramesh2021station}.

As reported in~\cite{makridakis2023statistical} that deep learning models perform better than most statistical and ML models in time series forecasting tasks, many deep learning-based time series forecasting models have been proposed. A dominant wave of short-term demand forecasting deep learning models applies deep neural networks, especially recurrent neural networks (RNNs) such as the variants of LSTM~\cite{hochreiter1997long} for taxi demand forecasting~\cite{rodrigues2019combining}, short-term passenger flow forecasting~\cite{zhang2020deep, liu2019deeppf, noursalehi2021dynamic, chang2021understanding, ke2017short}, and travel demand forecasting~\cite{guo2020residual},  More specifically, Rodrigues et al.~\cite{rodrigues2019combining} leverage the textual data as the exogenous information for the taxi demand forecasting. The textual data is encoded along with time series historical demand data as the input of LSTM. To address the short-term passenger flow prediction issue, Zhang et al.~\cite{zhang2020deep} propose a novel deep learning structure that combines residual network, graph convolutional network, and LSTM. Liu et al.~\cite{liu2019deeppf} integrate the external environmental factors, temporal dependencies, and spatial characteristics with RNN that enables the DL model to be highly flexible and extendable. Similarly, Noursalehi et al.~\cite{noursalehi2021dynamic} study the original-destination (OD) traffic demand predicting problem by proposing a deep learning architecture that involves three modules. Multi-resolution spatial features, auxiliary information, and temporal evolution of demand are encoded and processed by the three modules, respectively. Chang et al.~\cite{chang2021understanding} investigate the demand forecasting issue in bike-sharing systems by proposing an end-to-end deep learning architecture. Aiming at short-term passenger demand forecasting, Ke et al.~\cite{ke2017short} propose a fusion convolutional LSTM that involves convolutional and LSTM layers. The hybrid model structure can better capture the spatio-temporal features and correlations of explanatory variables. A similar work can be found in~\cite{guo2020residual}, where a residual spatio-temporal neural network is proposed for short-term demand forecasting issues. A hybrid module that consists of fully convolutional neural networks and Conv-LSTM is designed to capture the spatial, temporal, and extraneous dependencies of travel demand.

Recently, Transformer~\cite{vaswani2017attention} based models, which are based on the attention mechanism, have become a trend in short-term demand forecasting in MoD systems. Li et al.~\cite{li2021intercity} study an online car-hailing demand prediction problem by proposing a spatio-temporal transformer model. The spatio-temporal features are extracted and encoded into the transformer model, and historical demands of the divided areas are leveraged to predict future demand. Likewise, Bi et al.~\cite{bi2022spatiotemporal} divide the car-hailing region into several grids and convert the travel data into video frames by time period. The converted information is then fed into the transformer model for travel demand forecasting. Xu et al.~\cite{xu2023multi} propose a transformer-encoder-based neural process to address the supply-demand prediction problem in the bike-sharing system. The neural process model learns a Gaussian process by mapping the observation values to a probabilistic distribution over functions. Wen et al.~\cite{wen2023traffic} present a spatio-temporal guided multi-graph sandwich-transformer to address the traffic demand forecasting issue. Time to vector and node to vector are embedded to obtain the representations for both spatial and temporal nodes. Koohfar et al.~\cite{koohfar2023prediction} apply the transformer model to forecast both long-term and short-term electric vehicle charging demand issues. In addition, a group of graph neural network (GNN) models are also discussed in the literature. Liang et al.~\cite{liang2023cross} propose a domain-adversarial-based GNN for bike-sharing demand forecasting. Benefiting from the adapted multi-relational GNN, the GNN-based model can fuse spatio-temproal information effectively. Zhuang et al.~\cite{zhuang2022uncertainty} study an Origin-Destination (O-D) demand prediction issue in MoD systems, where a spatio-temporal GNN is designed to quantify the uncertainty of the sparse travel demand. Lin et al.~\cite{lin2018predicting} study a station-level hourly demand forecasting problem in bike-sharing systems through a tailored GCNN, where the hidden heterogeneous pairwise correlations can be learnt to forecast hourly demand in large-scale bike-sharing systems.

However, the deep learning models above are based on point prediction that ignores uncertainty. In addition, the DeepAR model~\cite{salinas2020deepar} is a state-of-the-art deep learning model for probabilistic forecasting. Unlike the deep learning forecasting models above, DeepAR approximates a distribution as the time series forecasting results. Nevertheless, DeepAR applies a single Gaussian or Student t's distribution as the forecasting output. Due to its symmetric characteristic, it may not be applied to describe the forecasting distribution of MoD demand with high volatility. To overcome the limitations, we propose a novel deep-learning approach to address the short-term probabilistic demand forecasting issue considering the demand volatility.

\section{The Extended Recurrent Mixture Density Networks for Probabilistic Demand Forecasting} \label{section:approach}
In this section, we elaborate on the mechanism of our approach. First, we briefly describe the short-term demand forecasting problem in MoD systems. Next, we show the details of the proposed XRMDN structure, including the notation, neural network organization, loss function, etc.

\subsection{Short-Term Probabilistic Demand Forecasting in MoD}

We consider a fairly general MoD demand probabilistic forecasting problem. We divide the on-mobility service area into a group of grids (also known as regions). During each short time period (e.g., 10 minutes), the MoD systems collect the customer requests and aggregate them as the demand at this time interval. As the MoD systems operate over time, the demand information is accumulated as the historical demand data. 

\begin{definition}[Time Series Rider Demand Sequence]
Let $D_{t}$ be the rider demand information at time $t$, which is a composite data denoted by a 2-tuple $<d_t, f_t>$. $d_t$ and $f_t$ denote the rider demand value (i.e., the aggregation of rider requests) and the exogenous characteristics, such as spatial, temporal, and meteorological features, respectively. Then, the time series rider demand sequence can be represented as $(D_1, D_2, \cdots, D_t)$.
\end{definition}

Given the time series rider demand sequence above, the forecasting demand is modeled as a function of the time-dependent historical demand sequence. In particular, the short-term probabilistic demand forecasting problem aims to estimate the \textit{conditional probability distribution} $P$ of the rider demand $\widehat{d}_{t+1}$ at the next time $t+1$, given the historical k-length (a.k.a. lookback window k) historical rider demand sequence ($D_{t \text{-} k+1}, D_{t \text{-} k+2}, \cdots, D_t$), which is formulated in the following equation.

\begin{align}\label{eq:1}
    P (\widehat{d}_{t+1} \mid D_{t \text{-} k+1}, D_{t \text{-} k+2}, \cdots, D_{t})
\end{align}


\subsection{Methodology - XRMDN}

XRMDN extends RMDN, which adopts the concept of mixture density networks (MDN)~\cite{bishop1994mixture}, by encoding exogenous information to improve forecasting performance, especially when the rider demand is highly volatile. Furthermore, the XRMDN model significantly differs from RMDN in terms of the network architecture. Specifically, the XRMDN model consists of three sub-recurrent neural networks, namely, (1) \textit{Weight Recurrent Neural Network} (WRNN), (2) \textit{Mean Recurrent Neural Network} (MRNN), and (3) \textit{Variance Recurrent Neural Network} (VRNN). The training process of XRMDN is summarized in Algorithm~\ref{alg:XRMDN training}.

Given the training epoch, during each iteration (i.e., batch), the historical demand sequence and the exogenous information are embedded as the input for the paralleled sub-recurrent neural networks, WRNN and MRNN. The output features are then concatenated in the successive hidden layers of WRNN and MRNN. Given $N$ Gaussian components and the demand sequence with lookback window $t$, the forecasting weight vector and mean vector at time $t+1$ can be denoted as $\widehat{\eta}_{t+1} = (\widehat{\eta}_{1, t+1}, \widehat{\eta}_{2, t+1}, \cdots, \widehat{\eta}_{N, t+1})$ and $\widehat{\mu}_{t+1} = (\widehat{\mu}_{1, t+1}, \widehat{\mu}_{2, t+1}, \cdots, \widehat{\mu}_{N, t+1})$, respectively. For simplicity, we assume the number of Gaussian components is 2 ($N=2$), the formulas of WRNN and MRNN are shown in Eq.~(\ref{formular:weight}) and Eq.~(\ref{formular:mean}), respectively. Notice that the softmax activation function must be applied in the output layer WRNN since the sum of the weights must be equal to one, which is shown in Eq.~(\ref{formular:softmax}).


\begin{equation}
\begin{aligned}\label{formular:weight}
\widehat{\eta}_{i, t+1} & = softmax \left(w_{i, 1}\left(W_{1,1} \eta_{t} + W_{1,0}\right) + \right. \\
& \left. \sum_{k=2}^K w_{i, k} \tanh \left(W_{k, 1} \eta_{t} + W_{k, 0}\right) + \right. \\
& \left. w_{i, K+1} \left(W_{K+1,1} \widehat{\eta}_{i, t} + W_{K+1,0}\right) + \right. \\
& \left. \sum_{k=K+2}^{2 K} w_{i, k} \tanh \left(W_{k, 1} \widehat{\eta}_{i, t} + W_{k, 0}\right) + w_{i, 0} \right)
\end{aligned}
\end{equation}

\begin{equation}
\begin{aligned}\label{formular:mean}
\widehat{\mu}_{i, t+1} & = dense \left(w_{i, 1}\left(W_{1,1} \mu_{t} + W_{1,0}\right) + \right. \\
& \left. \sum_{k=2}^K w_{i, k} \tanh \left(W_{k, 1} \mu_{t} + W_{k, 0}\right) + \right. \\
& \left. w_{i, K+1} \left(W_{K+1,1} \widehat{\mu}_{i, t} + W_{K+1,0}\right) + \right. \\
& \left. \sum_{k=K+2}^{2 K} w_{i, k} \tanh \left(W_{k, 1} \widehat{\mu}_{i, t} + W_{k, 0}\right) + w_{i, 0} \right)
\end{aligned}
\end{equation}

\begin{align}\label{formular:softmax}
    softmax(\widehat{\eta}_{t+1})_{i} = \frac{e^{\widehat{\eta}_{i, t+1}}}{\sum_{j=1}^{N} e^{\widehat{\eta}_{j, t+1}}}
\end{align}

Since the output of the WRNN and the MRNN can be considered as the expected forecasting demand at time $t+1$, the demand residue $e_{t+1}^{2}$, like mean square error, can be calculated by Eq.~(\ref{equation:error}), 

\begin{align}\label{equation:error}
e_{t+1}^{2} = (\EX(\widehat{\mu}_{t+1}) - d_{t+1})^2
\end{align}

Take the above demand residual as the partial input for the VRNN (the other part is recurrent state of the VRNN); the WRNN and the MRNN are concatenated with the VRNN as the entire XRMDN structure. Meanwhile, the predicted demand variance vector $\widehat{\sigma}_{t}^{2} = (\widehat{\sigma}_{1, t}^{2}, \widehat{\sigma}_{2, t}^{2}, \cdots, \widehat{\sigma}_{N, t}^{2})$ at time $t$\footnote{Namely, the output of the VRNN from the previous iteration} is applied to initialize the hidden states in the VRNN\footnote{During the first iteration, namely at time $t=1$, the demand variance is obtained from the training set since no predicted variance output}. Namely, the demand residue from the WRNN and the MRNN is combined with the output of the VRNN together as the input of the VRNN, whose formula is shown in Eq.~(\ref{formula:variance}). Notice that a particular activation function that returns a positive value must be applied for the output layer of the VRNN because the forecasting variance must be positive. Therefore, we design a novel activation function named Positive Exponential Linear Unit (PELU), which is based on the Exponential Linear Unit (ELU) activation function. The PELU and ELU functions are shown in Eq.~(\ref{equation:PELU}) and Eq.~(\ref{equation:ELU}), respectively.

\begin{align} \label{equation:PELU}
    PELU(z) = ELU(z) + 1 + \xi
\end{align}

\begin{equation} \label{equation:ELU}
ELU(z)=\left\{\begin{array}{cc}
z & z>0 \\
\alpha \cdot\left(e^z-1\right) & z<=0
\end{array}\right\}
\end{equation}

Finally, given the sequence of time series demand data and the Gaussian mixture model with $N$ components, the loss function of XRMDN is the negative log-likelihood function, which is shown in Eq.~(\ref{equation:loss-function}).

\begin{align}\label{equation:loss-function}
-\ell & =\sum_{t=1}^T-\log \sum_{i=1}^N \widehat{\eta}_{i, t} \phi\left({d}_{t}; \widehat{\mu}_{i, t}, \widehat{\sigma}_{i, t}^{2}\right) 
\end{align}

Herein, $\phi(\bullet)$ represents the Gaussian distribution whose parameters are determined by XRMDN prediction results, as discussed above.

After the XRMDN model is trained, given a rider demand sequence for the probabilistic prediction, the probabilistic forecasting results from XRMDN consist of three parts: weight vector $\widehat{\eta}_{t+1}$, mean vector $\widehat{\mu}_{t+1}$, and variance vector $\widehat{\sigma}_{t+1}^{2}$, which are the parameters of Gaussian mixture model.

\begin{equation}
\begin{aligned}\label{formula:variance}
\widehat{\sigma}_{i, t+1}^{2} & =PELU \left(w_{i, 1}\left(W_{1,1} e_t^2+W_{1,0}\right) + \right. \\
& \left. \sum_{k=2}^K w_{i, k} \tanh \left(W_{k, 1} e_t^2+W_{k, 0}\right) + \right. \\
& \left. w_{i, K+1}\left(W_{K+1,1} \widehat{\sigma}_{i, t}^2+W_{K+1,0}\right) + \right. \\
& \left. \sum_{k=K+2}^{2 K} w_{i, k} \tanh \left(W_{k, 1} \widehat{\sigma}_{i, t}^2+W_{k, 0}\right)+w_{i, 0}\right)
\end{aligned}
\end{equation}

\begin{algorithm}
    \SetAlgoLined 
	\caption{XRMDN Training Process} \label{alg:XRMDN training}
	\KwIn{time series rider demand sequence data}
	Initialize the weights for WRNN, MRNN, and VRNN \;
 
    \While{the stopping criteria is not satisfied}
    {
		\ForEach{batch of the demand sequence}
        {
            feed the data to WRNN and MRNN \;
		      \eIf{the first batch of data}
            {
			    initialize the hidden state of WRNN with $\frac{1}{N}$, where $N$ denotes the number of Gaussian components\;
                initialize the hidden state of MRNN with demand mean \;
		      }
            {
			    get the hidden state of WRNN from the output of WRNN in the previous step \;
                get the hidden state of MRNN from the output of MRNN in the previous step \;
		      }
            feed the hidden states to WRNN and MRNN, respectively \;
            compute the residual using Eq.~(\ref{equation:error}) \;

            feed the residual to VRNN \;
            \eIf{the first batch of data}
            {
			    initialize the hidden state of VRNN with the demand variance\;
		      }
            {
			    get the hidden state of VRNN from the output of VRNN in the previous step \;
		      }
        }
        minimize the loss function Eq.~(\ref{equation:loss-function}) and update the weights of WRNN, MRNN, and VRNN ;\
	}

	return 
\end{algorithm}



\begin{figure*}[htbp]
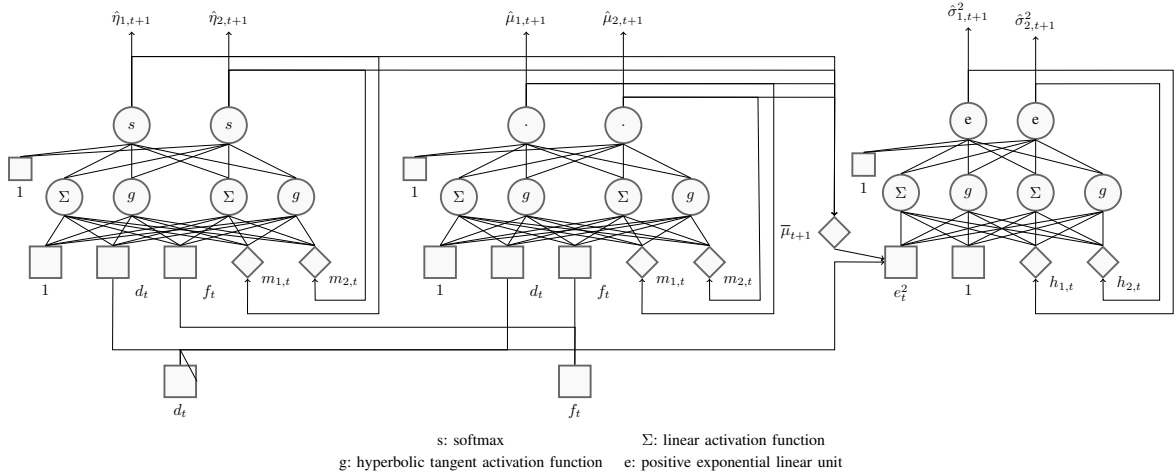

    \centering
    \include{rmdn_normal_architecture}
    \caption{The architecture of XRMDN. WRNN and MRNN are fed with demand sequence data in parallel, the residual is computed by the outputs of WRNN and MRNN, which is applied to concatenate VRNN. For simplicity, we assume there are only two Gaussian components in the architecture.}
    \label{fig:architecture}
\end{figure*}

The architecture of the XRMDN model is shown in Fig.~\ref{fig:architecture}. Like other types of deep neural networks, there should be multiple hidden layers for the WRNN, the MRNN, and the VRNN. For simplicity, we draw one hidden layer only to represent the basic architecture of the XRMDN model. In addition, we draw only two Gaussian components as the output for the WRNN, the MRNN, and the VRNN.

\section{Numerical Experiments} \label{section:experiment}

In this section, we conduct numerical experiments on two different real MoD datasets, the New York yellow taxi trip record~\footnote{https://www.nyc.gov/site/tlc/about/tlc-trip-record-data.page} and bike-sharing datasets from UCI~\footnote{https://archive.ics.uci.edu/dataset/275/bike+sharing+dataset}, to validate our approach. First, we describe the experiment setup and model parameters. Afterward, we specify data processing, feature selection, and data exploration. We then elaborate on three evolution metrics and three types of benchmark models. Finally, we compare the validation results and discuss the limitations of our work.

\subsection{Experiment Setup}
The deep learning forecasting models, including classic RMDN, XRMDN, and LSTM, are implemented using Python 3.9 \& PyTorch 1.13~\footnote{https://pytorch.org/}. DeepAR is implemented using GluonTS 0.13~\cite{gluonts_arxiv, garnelo2018neural}. All the deep learning models are run on a PC server with  GeForce RTX 2080 GPU, 16 GB RAM, and Ubuntu 18.04. The rest of the forecasting models are run on a PC with Intel Core i7, 32 GB RAM, and Windows 11. ARIMA and ARIMA-GARCH are implemented using statsmodels 0.13~\footnote{https://github.com/statsmodels/statsmodels}. The parameter settings of deep learning models are shown in Table~\ref{table:DL-parameters}.

\begin{table}[]
\centering
\caption{The parameter settings of deep learning models}
\small 
\setlength{\tabcolsep}{13.3pt}   
\renewcommand{\arraystretch}{1.5}  %
\begin{tabular}{ll}
\toprule[1pt]

{LSTM \& DeepAR~\footnotemark[1]}  &  {Settings} \\
\hline
Input layer & 3 and 9~\footnotemark[2]   \\  
Hidden layer  & 32  \\ 
Activation function & ReLU \\
Optimization algorithm & Adam~\cite{kingma2014adam} \\
Epoch & 50 \\

\toprule[1pt]

{RMDNs} & {Settings}\\ 
\hline
Input layer (XRMDN) & (5, 5, 3) and (11, 11, 3)~\footnotemark[3]   \\ 
Input layer (RMDN) & (3, 3, 3) and (9, 9, 3)~~\footnotemark[4]   \\ 
Hidden layer & (8, 8, 8)~~\footnotemark[5]  \\ 
Gaussian components & 2 \\
Activation function & Softmax, Tanh, PELU \\
Optimization algorithm & Adam \\
Epoch & 50 \\

\bottomrule[1pt]
\end{tabular}
\label{table:DL-parameters}
\end{table}
\footnotetext[1]{The output of the DeepAR model is set as a Student's t-distribution, which is the default option suggested by GluonTS. The rest of the parameters are identical to LSTM for the sake of fair comparison}
\footnotetext[2]{In LSTM, the number of neurons in the input layer is set to 3 (demand sequence and two temporal features) and 9 (demand sequence and eight features) to be applied to the New York dataset and bike-sharing dataset, respectively}
\footnotetext[3]{Unlike LSTM, in RMDNs (classic RMDN and XRMDN), the number of neurons is set in a different way. Take XRMDN on the New York dataset as an example, the numbers of neurons in the input layers are 5, 5, and 3. The explanations are described as follows. There are five neurons in the input layer for the WRNN and the MRNN (demand sequence, two temporal features, and two recurrent connections since the number of Gaussian components is set to two). Meanwhile, there are three neurons in the input layer for the VRNN (demand residual and two recurrent connections). By this mechanism, the numbers of neurons for the three sub-neural networks on the bike-sharing dataset are set to 11, 11, and 3, respectively}
\footnotetext[4]{Based on the analysis above, the number of neurons in the input layers of RMDN are identical to LSTM and DeepAR}
\footnotetext[5]{The number of neurons in the hidden layer of the WRNN, the MRNN, and the VRNN are all set to 8}

\subsection{Data Processing, Feature Selection, and Heteroscedasticity Analysis}

\subsubsection{\textbf{New York yellow taxi trip records dataset}}
\begin{figure}
    \centering
    \includegraphics[scale=0.5]{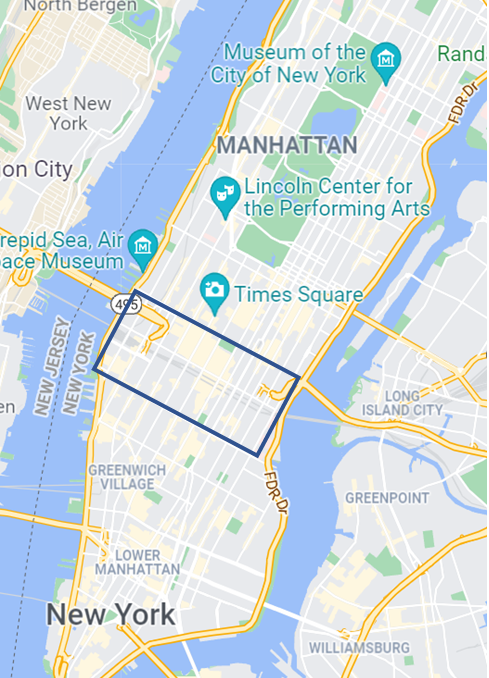}
    \caption{The selected region in Manhattan (NYC yellow taxi trip record dataset)}
    \label{fig:region}
\end{figure}

We use New York yellow taxi trip records in January (demand info in winter) and June 2016 (demand info in summer). There are millions of trip records and 19 columns in the dataset. For the validation purpose, we select one region as plotted in Fig.~\ref{fig:region}. In addition, the trip records are filtered and aggregated into 10 minutes (time interval) as the rider demands. Therefore, there are overall 144 values for daily time-dependent sequences. After the data processing, we split the entire data into training sets (January 1st - January 30th and June 1st - June 30th) and test sets (the last day of January and June), respectively. We identify two temporal features, including hour-of-day and day-of-week, for the deep learning models.

\subsubsection{\textbf{UCI Bike-sharing dataset}}

In this dataset, the rider demands are aggregated by day. There are two years of data (731 records) in total. The selected features from the dataset can be categorized into two types: the temporal features, including season, month, hour, and day-of-week, and the meteorological features, including temperature, feeling temperature, humidity, and wind speed. We split the dataset into a training set (January 2011 - August 2012) and a test set (September 2012 - December 2012).

To identify the heteroscedasticity existence in the datasets above, we leverage the processed training sets to fit an ARIMA model and obtain the forecasting residual $\xi$. We then apply the Ljung–Box test~\cite{box1970distribution} to explore the heteroscedasticity existence in the residual. The Ljung–Box statistics follow a $\chi_{\alpha, h}^2$ distribution, where $h$ (degrees of freedom) and $Q$ denote the number of lags and the value of Ljung–Box statistics, respectively. The null hypothesis ${H}_{0}$ assumes that there is no significant heteroscedasticity in the residual. The hypothesis testing results are displayed in Table~\ref{table:LJUNG-BOX TEST}, which implies high volatility exists in the time series demand sequences since the p-values are close to zero.

\begin{table}[]
\centering
\caption{The forecasting error $\xi$ on the training sets using the Ljung–Box test under significance level $\alpha=0.05$.}
\small 
\setlength{\tabcolsep}{6.5pt}   
\renewcommand{\arraystretch}{1.5}  %
\begin{tabular}{c | ccccc}
\toprule[1pt]
{Lags ($h$)} & {1} & {2} & {3} & {4} & {5}\\ 
{$\chi_{1-\alpha, h}^2$} & {3.842} & {5.992} & {7.815} & {9.488} & {11.071}\\ 
\hline
(DS-1) $Q$ & 0.15  & 10.66 & 92.28 & 186.91 & 320.96 \\ 
p-value & 6.9e-01  & 4.8e-03 & 7.1e-20 & 2.4e-39 & 3.1e-67 \\ 
\hline
(DS-2) $Q$ & 0.34  & 12.99 & 73.57 & 179.13 & 316.46 \\ 
p-value & 5.6e-01  & 1.5e-03 & 7.3e-16 & 1.1e-37 & 2.9e-66 \\ 
\hline
(DS-3) $Q$ & 23.22  & 24.88 & 33.69 &  46.26 & 50.82 \\ 
p-value & 1.4e-06  & 3.9e-06 & 2.3e-07 & 2.2e-09 & 9.4e-10  \\ 
\bottomrule[1pt]
\end{tabular}
\label{table:LJUNG-BOX TEST}
\end{table}

For simplicity of presentation, in the following experiment results, we use DS-1, DS-2, and DS-3 to denote the New York yellow taxi trip record in January, June, and the UCI bike-sharing dataset, respectively.

\subsection{Evaluation Metrics}
We apply three groups of metrics for the performance evaluation. We use Mean Absolute Percentage Error (MAPE), Mean Absolute Error (MAE), and Root Mean Square Error (RMSE) to evaluate the performance for the point estimation based on the expected value. In addition, we use Log-Likelihood Value (LLV) and user-defined Rejection Rate (RR) to evaluate the performance of probabilistic forecasting results.

\begin{itemize}
    \item LLV. LLV is a negative number that can be used to quantify between value and distribution. A higher LLV implies the model can better fit the test data. The LLVs can be computed from Eq.(\ref{equation:loss-function}) by returning the absolute values.
    \item MAPE, MAE, and RMSE are defined as below.
    \begin{equation}
    \begin{aligned}
    & MAPE = \frac{1}{T} \sum_{t=1}^T \frac{\left| {d}_{t} - \widehat{d}_{t} \right|}{d_t} \\
    & MAE = \frac{1}{T} \sum_{t=1}^T \left| d_{t} - \widehat{d}_{t} \right| \\
    & RMSE = \sqrt{\frac{1}{T} \sum_{t=1}^T\left({d}_{t} - \widehat{d}_{t} \right) ^ {2}}
    \end{aligned}
    \end{equation}
    \item RR. We introduce this metric since it is useful to measure the quality of probabilistic forecasting results. Specifically, a rejection test case means the true demand at time $t$ does not fall into the samples' confidence interval that is determined by the given percentile. Based on this definition, we define the rejection rate as a ratio of the total number of rejection test cases over the number of test cases. Therefore, the rejection rate can be computed from Eq.~(\ref{equation:rejection-rate}).
    \begin{equation}\label{equation:rejection-rate}
       RR(p) = \frac{\sum_{t=1}^{K} \mathds{1}_{A(p)}(d_t)}{K}
    \end{equation}
    
    where $p$ is the given percentile value, $d_t$ is the true demand at time $t$, and $K$ is the number of test case. In our experiment, we apply 75\%, 90\%, and 95\% percentiles for the comparison. Meanwhile, the values of $K$ are 144 and 122 on the New York yellow taxi trip record and bike-sharing datasets, respectively. In addition, $\mathds{1}_{A(q)}(d_t)$ is an indicator function, which is defined in Eq.~(\ref{equation:indicator-function}).
    \begin{equation} \label{equation:indicator-function}
    \mathds{1}_{A(p)}(d_t):= \begin{cases}1 & \text { if } d_t \in A(p) \\ 0 & \text { if } d_t \notin A(p)\end{cases}
    \end{equation} 

    This indicator function returns 1 if the true demand $d_t$ is in the set of samples $A(p)$ that is determined by the percentile $p$ (aka confidence interval) and returns 0 otherwise (i.e., the true demand does not fall within the confidence interval given by the percentile).
    
    \end{itemize}

\subsection{Benchmark Models}

Aligned with forecasting models taxonomy discussed in Sec.~\ref{section:literature}, we select the following three groups (i.e., statistical, machine learning, and deep learning models) of time series forecasting models as the benchmark competitors.

\subsubsection{\textbf{Statistical Benchmarks}}
We select ARIMA and ARIMA-GARCH as the statistical competitors. ARIMA is the most classic time series forecasting model that is used in most of the literature. Hence, we apply this model as one of our benchmarks. Notice that there are three parameters for a given ARIMA(p, d, q) model, namely, the order of AR p, the order of differencing d, and the order of MA q. The integer values of p, d, and q are set to range from 1 to 7, from 1 to 2, and from 1 to 5, respectively. We select the best ARIMA model with the combined parameters based on the Akaike Information Criterion (AIC). On the other hand, ARIMA-GARCH integrates ARIMA, which focuses on mean value forecasting, with GARCH, which aims to predict the residual that is determined by the previous residuals and variances. The reason for selecting this model is that it considers both means and variances in the time series forecasting result, which is quite close to the mechanism of XRMDN. The ARIMA-GARCH formulation can be described as follows.

\begin{align}\label{formular:GARCH-1}
\widehat{d}_t = c_{0} + \sum_{i=1}^p a_i d_{t-i} + \sum_{j=1}^q b_j \epsilon_{t-j} + \epsilon_t 
\end{align}

\begin{align}\label{formular:GARCH-2}
\epsilon_t = e_{t} \sqrt{\sigma_{t}^{2}} = e_t \sqrt{\EX \left(\epsilon_t^{2} \mid I_{t-1}\right)}
\end{align}

\begin{align}\label{formular:GARCH-3}
\sigma_{t}^{2}= \gamma_{0} + \sum_{i=1}^p \alpha_i \epsilon_{t-i}^2 + \sum_{j=1}^q \beta_j \sigma_{t-j}^2
\end{align}

Unlike ARIMA, which assumes the term $\epsilon_{t}$ in Eq.~(\ref{formular:GARCH-1}) is white noise, GARCH relaxes the assumption by treating $\epsilon_{t}$ as a conditional heteroskedasticity, which is shown in Eq.~(\ref{formular:GARCH-2}). $e_{t}$ and $I_{t-1}$ denote the independently and identically distributed (i.i.d) random variable with zero mean and unit variance at time $t$, and available demand information at time $t-1$, respectively. In our following experiments, the GARCH model is set with the order (p=1, q=1) since it is good enough to represent the high volatility for most of the applications as reported in~\cite{tsekeris2006real}.

\subsubsection{\textbf{Machine Learning Benchmarks}}
We adopt the Light Gradient Boosting Machine (LightGBM)~\cite{ke2017lightgbm}, an ensemble learning model based on the gradient boosting technique. The reason we choose LightGBM is that it can achieve a better performance in MoD time series demand prediction compared to other types of machine learning models~\cite{chuwang2023short, liu2019spatio}.

\subsubsection{\textbf{Deep Learning Benchmarks}}
There are a large number of deep learning time series forecasting models. Based on the forecasting result forms, we select LSTM, which is the classic time series point prediction model, DeepAR, and classic RMDN, which are probabilistic prediction models, as the benchmarks~\footnote{Notice that there is a number of deep learning models for MoD demand forecasting, which integrates the convolutional neural networks (CNN). These types of models have been discussed in Section~\ref{section:literature}. However, image data is required when applying these approaches. Since we leverage textual and numerical information only in XRMDN, we do not compare the performance between XRMDN to the CNN-like deep learning approaches}.

\subsection{Experiment Results and Analysis}

\subsubsection{\textbf{Log-likelihood values comparison}}
\begin{figure*}
\resizebox{\textwidth}{!}{
\begin{minipage}[t]{0.3\linewidth}
\centering
\subfloat[]{
\includegraphics[width=5.5cm]{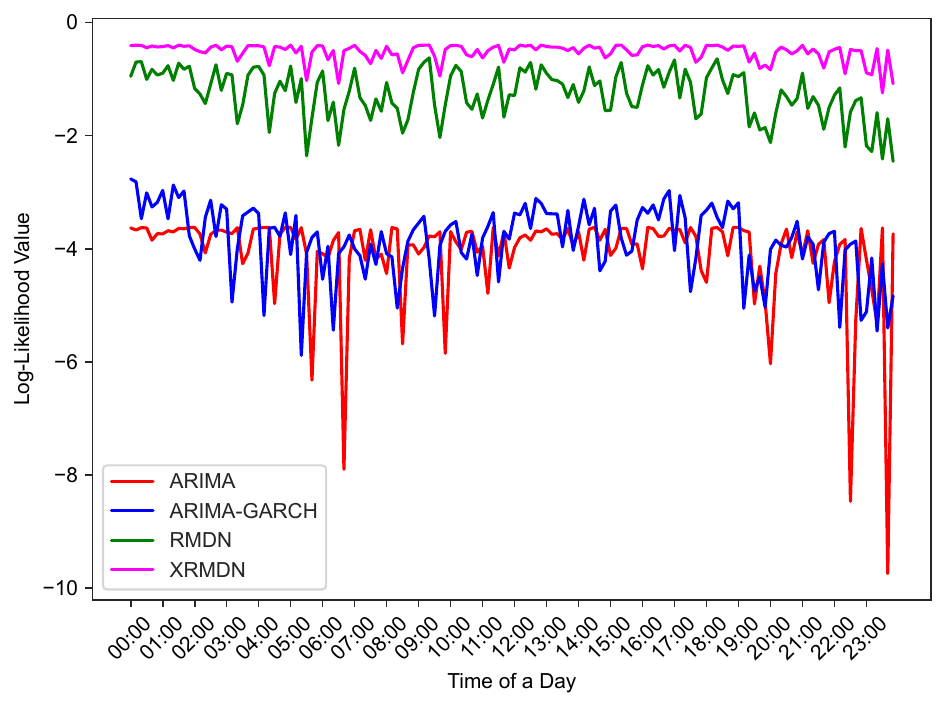}}
\end{minipage}

\begin{minipage}[t]{0.3\linewidth}
\centering
\subfloat[]{
\includegraphics[width=5.5cm]{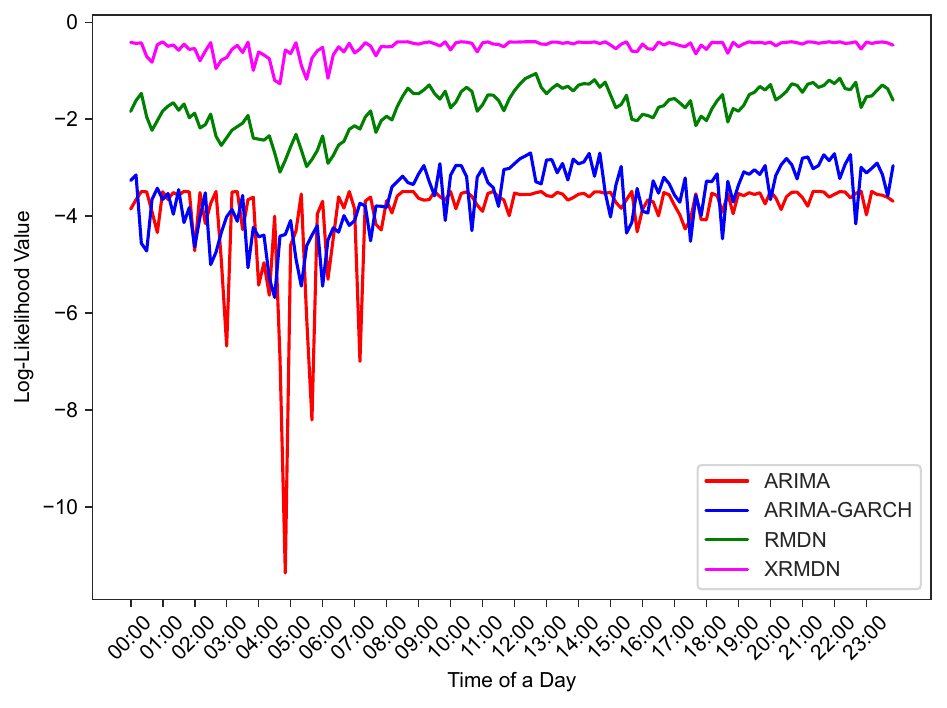}}
\end{minipage}

\begin{minipage}[t]{0.3\linewidth}
\centering
\subfloat[]{
\includegraphics[width=5.5cm]{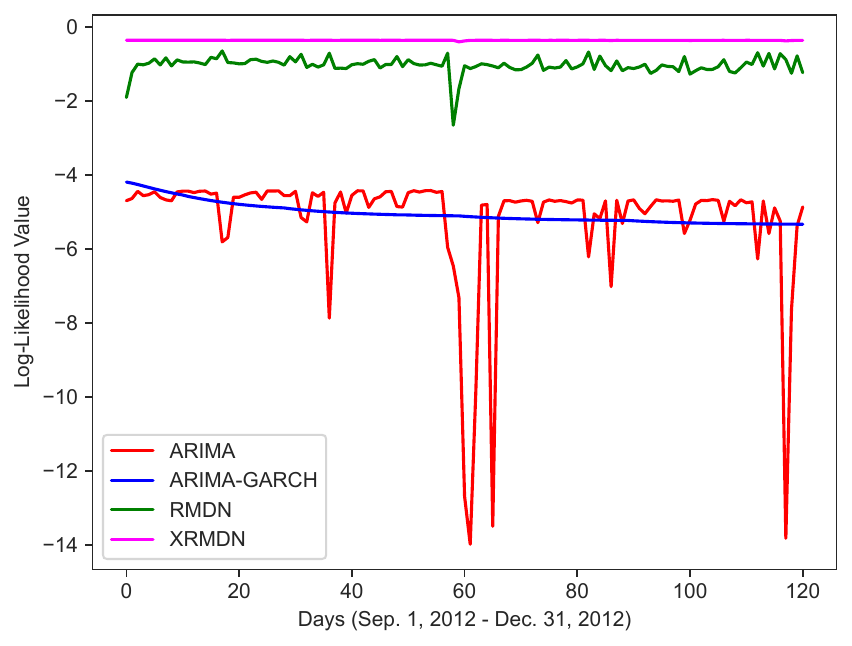}}
\end{minipage}}
\caption{Log-likelihood value comparison of ARIMA, ARIMA-GARCH, RMDN, and XRMDN on the three test sets, (a) DS-1, (b) DS-2, and (C) DS-3}
\label{fig:LLV comparison}
\end{figure*}

First, we compare the log-likelihood (LLV) of XRMDN to regular RMDN, ARIMA, and ARIMA-GARCH. The single LLV and overall LLVs comparison results are shown in Fig.~\ref{fig:LLV comparison} and Table~\ref{table:total-log-likelihood}, respectively. 

In general, the deep learning models (XRMDN and RMDN) are superior to the statistical models (ARIMA and ARIMA-GARCH) in terms of LLV across the two datasets. The experimental results are consistent with the conclusion as reported in~\cite{makridakis2023statistical}. In addition, it can be observed that the LLV from the ARIMA models drops dramatically when the true demands decrease with high volatility, as shown in Fig.~\ref{fig:demand-comparison}. While XRMDN, RMDN, and  ARIMA-GARCH are more robust to handle the demand with high volatility, as the forecasting result curves are stable across the test sets, especially on the bike-sharing dataset. This is because, for demand forecasting, the ARIMA model leverages the previous time-dependent demand sequence only while neglecting the variance effect on the forecasting results. Hence, ARIMA fails to capture the trend when demands drop with high volatility. On the contrary, XRMDN, RMDN, and ARIMA-GARCH leverage the variance information for demand forecasting. XRMDN and RMDN utilize the VRNN for the demand calibration, while ARIMA-GARCH utilizes the GARCH part for the demand forecasting calibration. Therefore, the forecasting results from the three models are robust and smooth. 

Furthermore, XRMDN performs best among the models as indicated in Table~\ref{table:total-log-likelihood}. Specifically, in terms of LLV, XRMDN is around 7.7 and 2.9 times higher than ARIMA and RMDN on average, respectively, on the New York dataset. The corresponding values are 14.2 and 2.8 on the bike-sharing dataset. Compared to the classic RMDN, XRMDN improves the network architecture by adding weight and mean recurrent connections. Hence, XRMDN can better handle the demand with high volatility. In addition, the LLV gap between XRMDN and ARIMA increases dramatically as the demand level becomes higher~\footnote{The demand on the bike-sharing dataset is over ten times higher than the demand on the New York dataset}, from 7.7 to 14.2. While the LLV gap between XRMDN and RMDN remains practically unchanged, from 2.9 to 2.8. The results suggest that compared to ARIMA and ARIMA-GARCH, XRMDN and RMDN are less sensitive to the demand level.

\begin{table}[]
\centering
\caption{The overall LLV comparison on the three test sets. The highest LLV indicate the best-fitting models, which are highlighted in bold font}
\small 
\setlength{\tabcolsep}{5.pt}   
\renewcommand{\arraystretch}{1.5}  %
\begin{tabular}{ccccc}
\toprule[1pt]
{} & {ARIMA} & {ARIMA-GARCH} & {RMDN} & {XRMDN}\\ 
\hline
LLV (DS-1) & -584.8  & -552.8 & -180.5 & \textbf{-74.9}\\ 
LLV (DS-2) & -563.5  & -515.2 & -253.2 & \textbf{-73.3}\\ 
LLV (DS-3) & -629.2  & -610.4 & -125.3 & \textbf{-44.2}\\ 
\bottomrule[1pt]
\end{tabular}
\label{table:total-log-likelihood}
\end{table}

\subsubsection{\textbf{Comparison between true and forecasting demands}}
\begin{figure*}
\resizebox{\textwidth}{!}{
\begin{minipage}[t]{0.3\linewidth}
\centering
\subfloat[]{
\includegraphics[width=5.5cm]{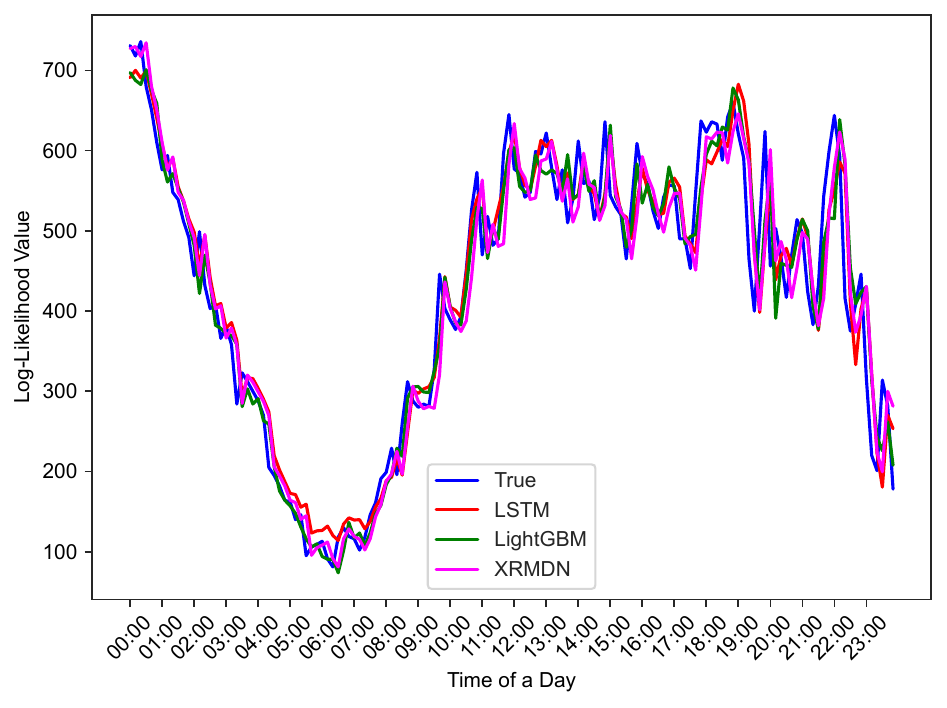}}
\end{minipage}

\begin{minipage}[t]{0.3\linewidth}
\centering
\subfloat[]{
\includegraphics[width=5.5cm]{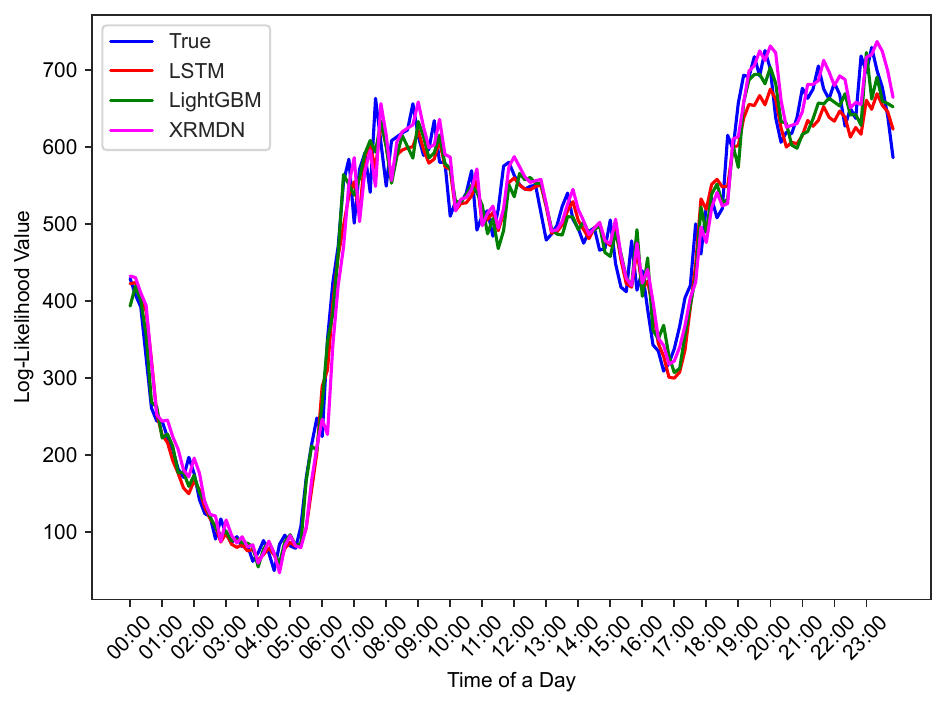}}
\end{minipage}

\begin{minipage}[t]{0.3\linewidth}
\centering
\subfloat[]{
\includegraphics[width=5.5cm]{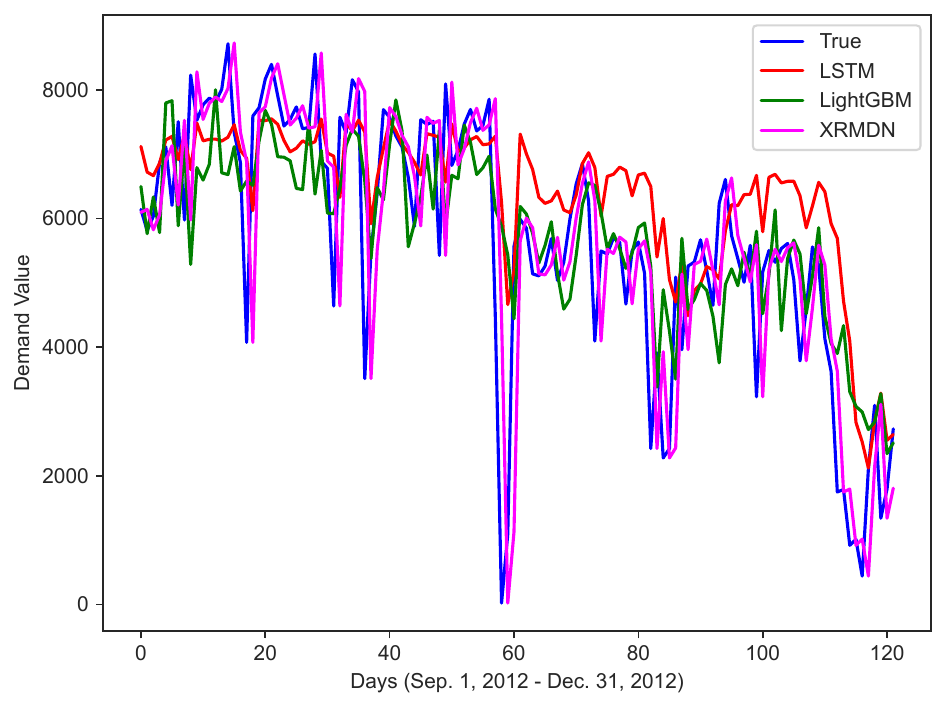}}
\end{minipage}}
\caption{Comparison of point forecasting demand results of LSTM, LightGBM, and XRMDN with the observed demand on the three test sets, (a) DS-1, (b) DS-2, and (C) DS-3}
\label{fig:demand-comparison}
\end{figure*}

second, we compare the XRMDN to XGBoost, LSTM, and Transformer from the perspective of point prediction results in terms of the three criteria: MAPE, MAE, and RMSE. Since XRMDN outputs the GMM parameters ($\eta$, $\mu$, $\sigma$) as the forecasting results, we set $\widehat \mu = \sum_{i=1}^{k} \eta_{i}\mu_{i}$, where $k$ denotes the number of Gaussian components, as the point forecasting results. The forecasting demands comparisons are shown in Fig.~\ref{fig:demand-comparison}. The comparisons of the three criteria are listed in Table~\ref{table:MSE-MAE-MAPE}.

Generally speaking, there are no significant differences among the three models in terms of the evaluation metrics on the New York trip record datasets. LightGBM is slightly better than XRMDN, while the LSTM model is relatively inferior to the LightGBM model and the XRMDN model. However, XRMDN considerably outperforms LightGBM and LSTM on the bike-sharing dataset. Specifically, there are around ten days when the demands are highly volatile on the bike-sharing dataset, as can be seen from the last plot in Fig.~\ref{fig:demand-comparison}. In particular, XRMDN can perfectly handle these scenarios with more accurate demand forecasting results. On the contrary, both LSTM and LightGBM fail to predict the demands of these days, which leads to higher RMSE, MAE, and MAPE values. Furthermore, in terms of MAE, XRMDN reduces the forecasting errors on the bike-sharing test set by 8.3\% and 18.3\% compared to LightGBM and LSTM, respectively. In terms of MAPE, the corresponding values are 25.7\% and 30.5\%. This is because XRMDN can better capture the demand trend with high volatility. Consequently, we can observe that the forecasting values by XRMDN are quite close to the real demands.

\begin{table}[]
\centering
\caption{MSE, MAE, and MAPE comparison on the three test sets. The lowest values on each row are highlighted in bold font}
\small 
\setlength{\tabcolsep}{10.5pt}   
\renewcommand{\arraystretch}{1.5}  
\begin{tabular}{cc|ccc}
\toprule[1pt]
\multicolumn{2}{c|}{} & {LSTM} & {LightGBM} & {XRMDN}\\ 
\hline
     & DS-1 & 53.9  & \textbf{46.5} & 47.6 \\ 
RMSE & DS-2 & 37.6  & \textbf{34.1} & 36.3 \\ 
     & DS-3 & 1527.9  & 1360.3 & \textbf{1320.4} \\ 
     \hline
     & DS-1 & 43.6  & \textbf{34.3} & \textbf{34.3} \\ 
MAE  & DS-2 & 29.9  & \textbf{26.0} &  28.8 \\ 
     & DS-3 & 1106.8 & 986.3 & \textbf{904.1} \\ 
     \hline
     & DS-1 & 0.15  & \textbf{0.09} & 0.10 \\ 
MAPE & DS-2 & 0.08  & \textbf{0.07} & 0.08 \\ 
     & DS-3 & 2.66  & 2.49 & \textbf{1.85}\\

\bottomrule[1pt]
\end{tabular}
\label{table:MSE-MAE-MAPE}
\end{table}

\subsubsection{\textbf{Comparison of rejection rates under different percentiles}}

\begin{figure*}
\resizebox{\textwidth}{!}{
\begin{minipage}[t]{0.3\linewidth}
\centering
\subfloat[]{
\includegraphics[width=5.5cm]{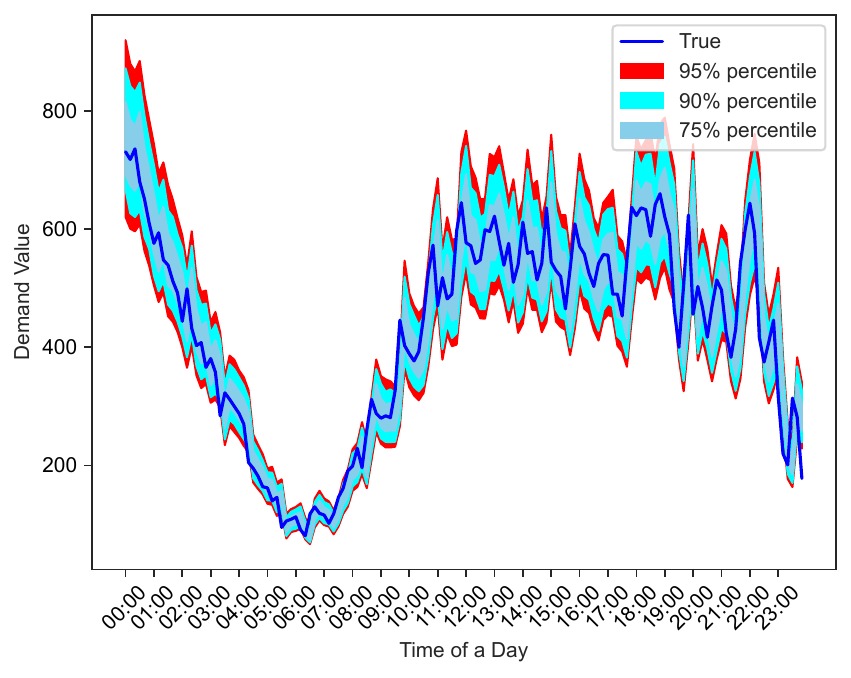}}
\end{minipage}

\begin{minipage}[t]{0.3\linewidth}
\centering
\subfloat[]{
\includegraphics[width=5.5cm]{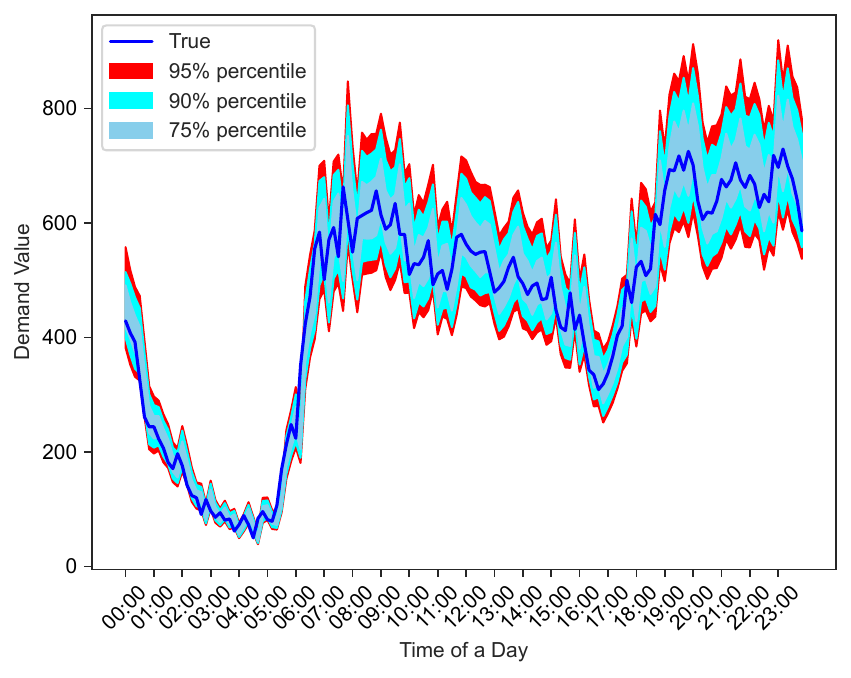}}
\end{minipage}

\begin{minipage}[t]{0.3\linewidth}
\centering
\subfloat[]{
\includegraphics[width=5.5cm]{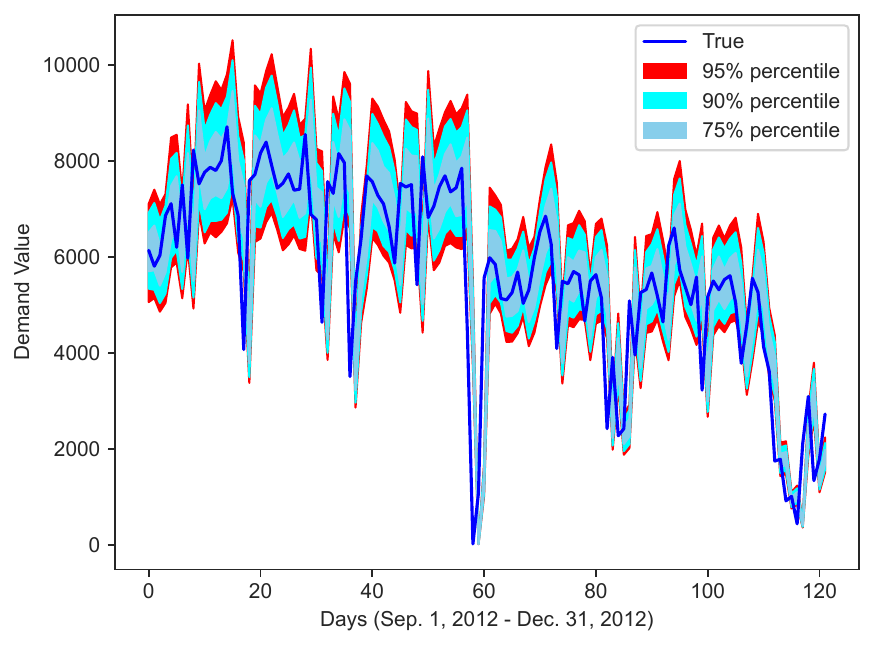}}
\end{minipage}}
\caption{The sample outcomes from XRMDN under different percentiles on the three test sets, (a) DS-1, (b) DS-2, and (C) DS-3}
\label{fig:XRMDN-sampling-comparison}
\end{figure*}

\begin{figure*}
\resizebox{\textwidth}{!}{
\begin{minipage}[t]{0.3\linewidth}
\centering
\subfloat[]{
\includegraphics[width=5.5cm]{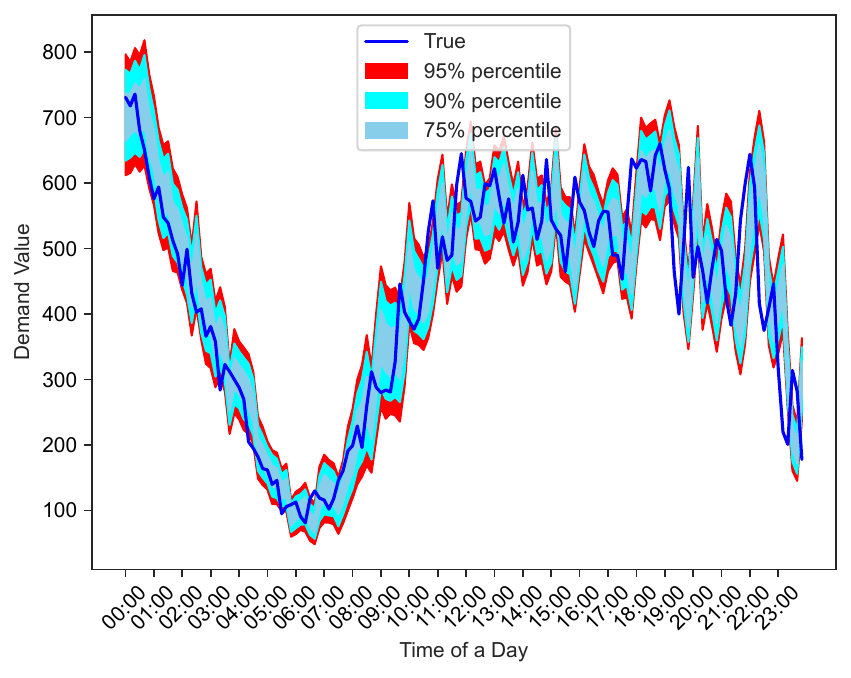}}
\end{minipage}

\begin{minipage}[t]{0.3\linewidth}
\centering
\subfloat[]{
\includegraphics[width=5.5cm]{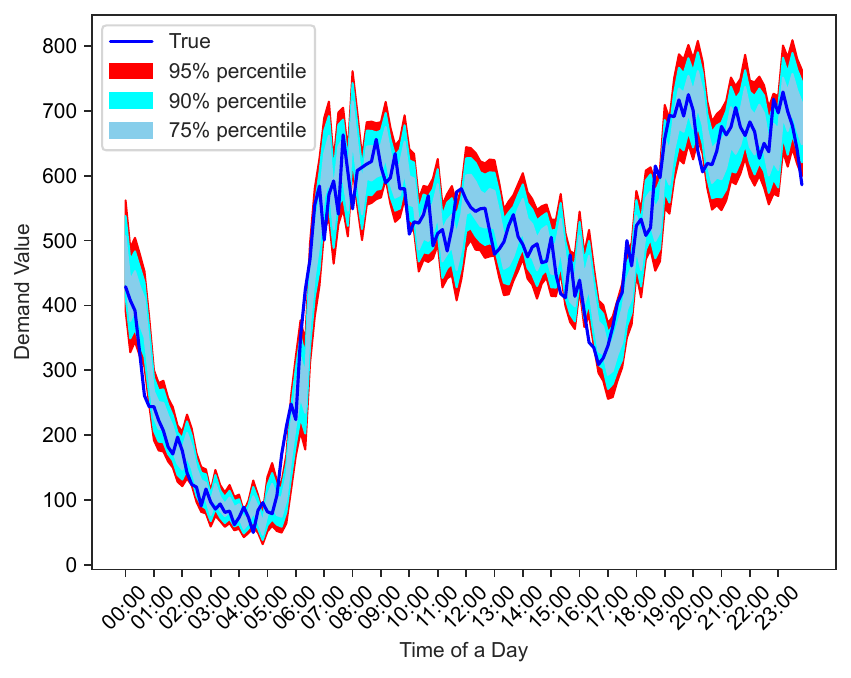}}
\end{minipage}

\begin{minipage}[t]{0.3\linewidth}
\centering
\subfloat[]{
\includegraphics[width=5.5cm]{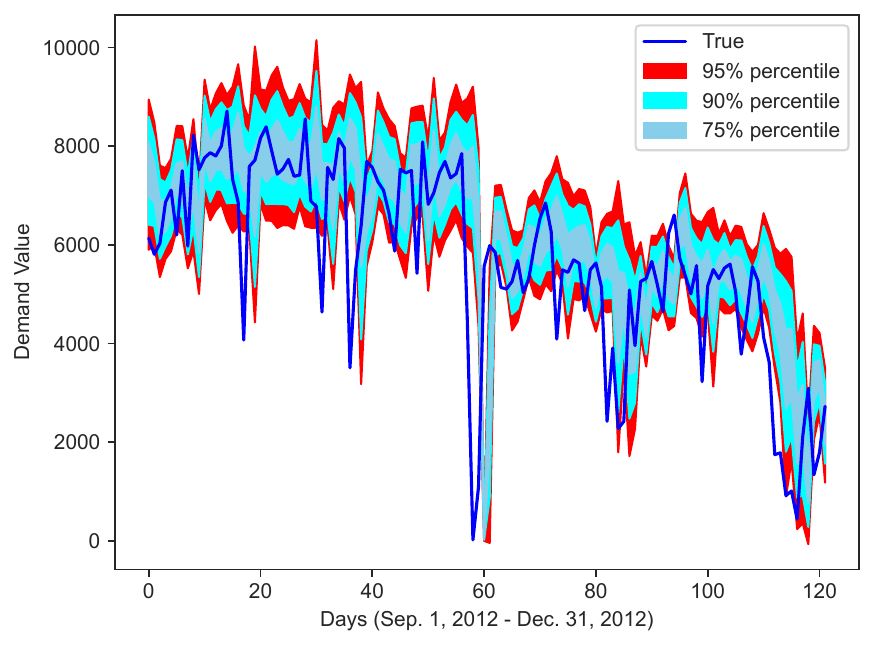}}
\end{minipage}}
\caption{The sample outcomes from DeepAR under different percentiles on the three test sets, (a) DS-1, (b) DS-2, and (C) DS-3}
\label{fig:DeepAR-sampling-comparison}
\end{figure*}

Finally, we compare the XRMDN to DeepAR in terms of rejection rate under different percentiles. Specifically, we draw 1,000 samples from each predicted distribution by XRMDN and DeepAR and compare the value ranges under 75\%, 90\%, and 95\% percentiles. The comparative results are shown in Fig.~\ref{fig:XRMDN-sampling-comparison} and Fig.~\ref{fig:DeepAR-sampling-comparison}. 

In general, it is observed that the probabilistic forecasting results from XRMDN provide a tight confidence interval compared to DeepAR, especially on the bike-sharing test set. Benefiting from the recurrent variance, XRMDN is capable of capturing the high volatility. Further, we observe that DeepAR also fails to yield high-quality forecasting results but is better than point forecasting models on the bike-sharing dataset when the demand is in high volatility. 

In addition, we compare the rejection rates between XRMDN and DeepAR under different percentiles on the three test sets. The comparative results are shown in Table~\ref{table:CI-rejection}. We observe that XRMDN substantially outperforms DeepAR in terms of the rejection rate on the New York test sets. This trend is consistent across the three percentile levels on the New York test sets. Compared to DeepAR, XRMDN can reduce the rejection rate by 45.0\% on average, up to 63.1\%. Moreover, we observe that XRMDN is slightly better than DeepAR on the bike-sharing test set except when the percentile is equal to 95\%. However, DeepAR reaches the goal by providing a much wider confidence interval. In fact, we can see that there exist a few exceptional points from the sampling results by the DeepAR model. For example, several sampling values on day-61 and day-118 are negative values, as can be seen from the last plot in Fig.~\ref{fig:XRMDN-sampling-comparison}. This implies that the variance in the forecasting result from the DeepAR model would be too large. To sum up, we come to conclude that the rejection rate is significantly reduced by XRMDN, which indicates that the probabilistic forecasting result from XRMDN can better approximate the true distribution on the test sets.

\begin{table}[]
\centering
\caption{The rejection rate comparison on the three test sets under 95\%, 90\%, and 75\% percentiles. The lower rejection rates are highlighted in bold font}
\small 
\setlength{\tabcolsep}{13.9pt}   
\renewcommand{\arraystretch}{1.5}  %
\begin{tabular}{cccc}
\toprule[1pt]
{} & {DS-1} & {DS-2} & {DS-3} \\ 
\hline
XRMDN (95\%) & \textbf{13.9\%}  & \textbf{9.0\%} & 33.6\% \\ 
DeepAR (95\%) & 27.1\%  & 15.2\% & \textbf{29.5\%} \\ 
\hline
XRMDN (90\%) & \textbf{20.2\%}  & \textbf{17.4\%} & \textbf{38.5\%} \\ 
DeepAR (90\%) & 39.6\%  & 29.2\% & 41.0\% \\ 
\hline
XRMDN (75\%) & \textbf{52.8\%}  & \textbf{38.2\%} & \textbf{54.9\%} \\ 
DeepAR (75\%) & 64.6\%  & 63.2\% & 66.4\% \\ 
\bottomrule[1pt]
\end{tabular}
\label{table:CI-rejection}
\end{table}

\subsection{Limitations and Discussion}
In this section, we elucidate two principal limitations inherent in the XRMDN framework. First, the architecture of XRMDN, integrating parallel configurations of WRNN, MRNN, and VRNN, engenders a complexity surpassing other deep learning models, such as DeepAR. This complexity increment leads to a more protracted training duration. Empirical observations, under identical parameter settings as delineated in Table~\ref{table:DL-parameters}, reveal that XRMDN's training duration is approximately 2.3 times more than that of LSTM. Second, certain temporal intervals may be more aptly characterized by a singular Gaussian or Student's t-distribution rather than a complex mixture model. In such instances, XRMDN's application might precipitate overfitting or yield imprecise forecasting outcomes.

\section{Conclusions and Future Work} \label{section:conclusion}

We introduce XRMDN, a novel deep learning framework designed to accurately forecast probabilistic demand in high-volatility Mobility-on-Demand (MoD) systems. This innovative model enhances the traditional RMDN by integrating recurrent neurons and encoding exogenous information, facilitating nuanced time series probabilistic predictions. By employing a Gaussian mixture model, XRMDN adeptly approximates the anticipated demand distribution. Empirical evaluations, utilizing authentic MoD datasets, affirm XRMDN's superior predictive performance over established statistical, machine learning, and contemporary deep learning methodologies in terms of key evaluation metrics.

Our future research will explore two promising avenues. First, recognizing the substantial improvement in forecasting accuracy under volatile demand scenarios afforded by GARCH models over traditional ARIMA, we propose the integration of GARCH methodologies within XRMDN. This endeavor seeks to harmonize the precision of statistical models with the predictive power of deep learning technologies. Second, acknowledging that XRMDN produces demand distributions as outputs, we aim to merge these forecasts with stochastic optimization models.


%



\bibliography{ref}
\bibliographystyle{IEEEtran}

\vfill

\end{document}